\documentclass[letterpaper]{article} 
\usepackage{aaai25}  
\usepackage{times}  
\usepackage{helvet}  
\usepackage{courier}  
\usepackage[hyphens]{url}  
\usepackage{graphicx} 
\usepackage{natbib}  
\usepackage{caption} 
\usepackage{multirow}
\usepackage{threeparttable}
\usepackage{arydshln}
\usepackage{booktabs}
\usepackage{algorithm}
\usepackage{algorithmic}

\usepackage{newfloat}
\usepackage{listings}
\DeclareCaptionStyle{ruled}{labelfont=normalfont,labelsep=colon,strut=off} 
\floatstyle{ruled}
\newfloat{listing}{tb}{lst}{}
\floatname{listing}{Listing}
\title{DABL: Detecting Semantic Anomalies in Business Processes Using Large Language Models}
\author{
    Wei Guan\textsuperscript{\rm 1}, 
    Jian Cao\textsuperscript{\rm 1}\thanks{Corresponding author.},
    Jianqi Gao\textsuperscript{\rm 1},
    Haiyan Zhao\textsuperscript{\rm 2},
    Shiyou Qian\textsuperscript{\rm 1}
}
\affiliations{
    \textsuperscript{\rm 1} Department of Computer Science and Engineering, Shanghai Jiao Tong University, Shanghai,  China\\
     \textsuperscript{\rm 2} Department of Computer Science and Engineering, University of Shanghai for Science and Technology, Shanghai, China

    \{guan-wei, cao-jian, 193139, qshiyou\}@sjtu.edu.cn, zhaohaiyan1992@foxmail.com
}

\usepackage{bibentry}

\begin{document}

\maketitle

\begin{abstract}
Detecting anomalies in business processes is crucial for ensuring operational success. While many existing methods rely on statistical frequency to detect anomalies, it's important to note that infrequent behavior doesn't necessarily imply undesirability. To address this challenge, detecting anomalies from a semantic viewpoint proves to be a more effective approach. However, current semantic anomaly detection methods  treat a trace (i.e., process instance) as multiple event pairs, disrupting long-distance dependencies.
In this paper, we introduce DABL, a novel approach for detecting semantic anomalies in business processes using large language models (LLMs). 
We collect 143,137 real-world process models from various domains. By generating normal traces through the playout of these process models and simulating both ordering and exclusion anomalies, we fine-tune Llama 2 using the resulting log.
Through extensive experiments, we demonstrate that DABL surpasses existing state-of-the-art semantic anomaly detection methods in terms of both generalization ability and learning of given processes.
Users can directly apply DABL to detect semantic anomalies in their own datasets without the need for additional training.
Furthermore, DABL offers the ability to interpret anomalies' causes in natural language, providing valuable insights into the detected anomalies.
\end{abstract}

%
\begin{links}
    \link{Code}{https://github.com/guanwei49/DABL}
\end{links}

\section{Introduction}
Business process anomaly detection is geared towards identifying undesired behavior occurring during process execution, serving as a crucial component in guaranteeing the efficient and dependable operation of businesses. By pinpointing anomalies within business processes, these detection techniques facilitate timely intervention, maintenance, and optimization, consequently bolstering overall well-being \cite{10738505}.

\begin{figure*}[tb]
	\centering
    \includegraphics[scale=0.27]{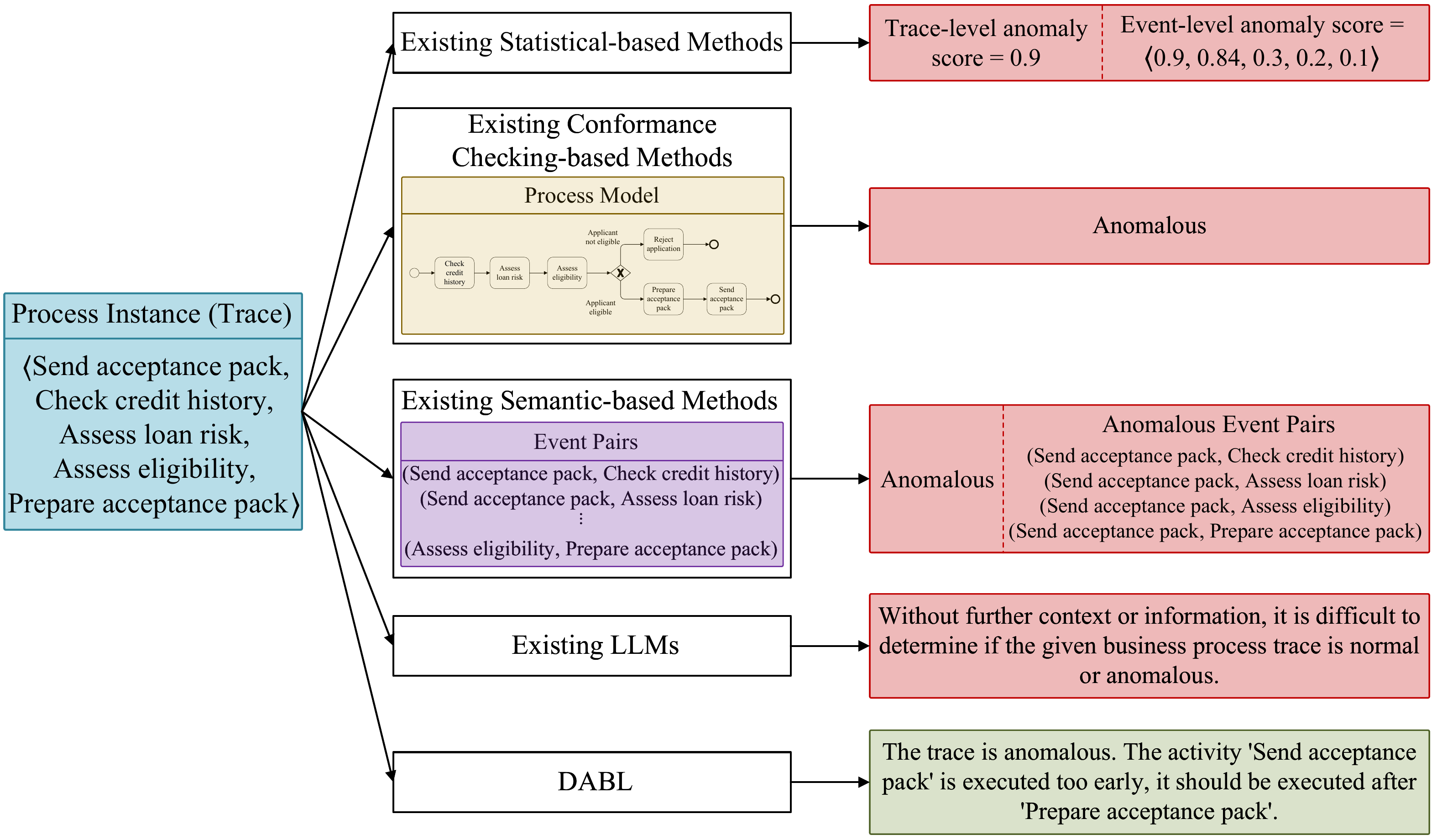}
	\caption{Comparison between our DABL with existing methods.}
	\label{fig:intro}
    \vskip -0.1in
\end{figure*}

Over the past few decades, notable advancements have been achieved in business process anomaly detection. Fig. \ref{fig:intro} illustrates a comparison of various applicable methods for accomplishing this task. 
Traditional statistical-based approaches \cite{Lu2022,Ko2022,nolle2022binet,Guan2023} rely on analyzing statistical frequencies to identify anomalies. However, infrequent behavior is not necessarily anomalous; it may represent rare but acceptable behavior. Conversely, frequent behavior may not always be normal. Furthermore, these methods focus on providing anomaly scores and require manual specification of thresholds to distinguish between normal and anomalous instances, which is not suitable for real-world applications.
Alternatively, methods based on conformance checking \cite{DBLP:journals/virology/EbrahimG22,DBLP:journals/jbd/SarnoSS20,sinaga2016business} detect anomalies by assessing the alignment between traces and their corresponding process models. Yet, accurately capturing complex processes within a process model remains a challenge, thereby restricting the utility of such approaches.
The concept of \textbf{semantic anomaly detection}, recently introduced, addresses these challenges by identifying anomalies from a semantic viewpoint. For example, it can detect irregularities such as a claim being paid after having been rejected.
Its grounding in natural language analysis allows for the consideration of typical behavior in standard processes, eliminating the necessity of having a specific process model at hand.
However, existing semantic-based anomaly detection methods \cite{van2021natural,caspary2023does} treat a trace as multiple event pairs, disrupting long-distance dependencies and thus limiting their accuracy.
Additionally, these methods interpret the cause of anomalies by providing anomalous event pairs, which can be confusing.

Recently, there have been significant advancements in LLMs. Due to their remarkable language comprehension abilities, LLMs such as GPT-3.5 \cite{ouyang2022training}, GPT-4 \cite{achiam2023gpt}, Llama 2 \cite{touvron2023llama}, and GLM-3 \cite{zeng2022glm} have shown proficiency in tasks like summarization, paraphrasing, and instruction following in zero-shot scenarios. However, in the context of semantic anomaly detection in business processes, their performance is limited by a lack of prior domain knowledge. As illustrated in Fig. \ref{fig:intro}, they often struggle to provide specific answers.

To address the aforementioned issues, we propose DABL, a fine-tuned LLM designed to detect semantic anomalies in business processes. 
Due to the lack of event logs comprising traces from various domains with rich semantic information, we generated our training dataset by playing out 143,137 real-world process models from three different process model datasets. This resulted in 1,574,381 normal traces. The collected process models cover a broad range of domains, including common processes related to order and request handling, as well as specialized processes in fields such as software engineering and healthcare.
Utilizing the generated normal traces, we then created synthetic anomalous traces. We introduced ordering anomalies, where activities should be executed in a different sequence (e.g., "accept request" followed by "check request"), and exclusion anomalies, where certain activities should not occur together within the same trace without an intermediate activity (e.g., "refusing the application" followed by "accepting the application" without "reapplying" in between). These generated normal and anomalous traces collectively form the training dataset.
Finally, by incorporating traces into question and answer content, we fine-tune the Llama 2-Chat 13B model \cite{touvron2023llama}, an open-source LLM, using QLoRA \cite{dettmers2024qlora}, to create a generic model capable of detecting semantic anomalies in business processes.
Compared to existing anomaly detection methods, DABL offers the capability to interpret the causes of anomalies in natural language, providing valuable insights into the detected anomalies. 
Extensive experiments show that DABL surpasses state-of-the-art methods in both generalization ability and learning of given processes.

Thanks to DABL's strong generalization ability, users can apply our open-source, fine-tuned model directly to their datasets without the need for additional training. Notably, it operates in a zero-shot manner, meaning it does not need normal traces or a process model during its operation.
Our contributions are summarized as follows:
\begin{itemize}
    \item We introduce DABL, an innovative method for fine-tuning LLMs to detect semantic anomalies in business processes. 
    \item We introduce novel techniques for simulating business process anomalies, encompassing ordering anomalies and exclusion anomalies, thereby enabling precise fine-tuning of LLMs.
    \item Extensive  experiments show  that DABL outperforms state-of-the-art methods in both generalization ability and the learning of given processes. Tests on real-world datasets confirm the practical effectiveness of DABL.
\end{itemize}

\section{Related Work}
\subsection{Business Process Anomaly Detection}
Existing business process anomaly detection methods can be divided into three categories: statistical-based, conformance checking-based, and semantic-based.

\subsubsection{Statistical-based Methods}
Some of these methods construct probabilistic models to infer the probability values (i.e., anomaly score) of traces. For example, HPDTMC \cite{9408048} constructs discrete-time Markov chains (DTMC) and introduces hitting probabilities (HP). EDBN \cite{10.1145/3297280.3297326, 10.1145/3357385.3357387} extends dynamic Bayesian networks by adding functional dependencies. PN-BBN \cite{Lu2022} extends Petri nets by incorporating Bayesian networks.
Other methods convert traces into vector representations and then detect anomalies using data mining techniques such as local outlier factor (LOF) and isolation forest (IF). For example, activities are considered as words and encoded into vectors using the word2vec \cite{DBLP:journals/corr/abs-1301-3781} technique in \cite{9230308, 10.1145/3466933.3466979}. Additionally, Trace2vec \cite{10.1007/978-3-319-98648-7_18}, extended from doc2vec and n-gram, is used to encode traces in \cite{rullo2020framework}.
Furthermore, the authors in \cite{Ko2022, Ko202153} employ one-hot encoding to convert traces into vector representations and detect anomalies using statistical leverage \cite{hoaglin1978hat}.
Recently, deep learning has been adopted to detect anomalies based on reconstruction errors. Given that traces exhibit sequential data characteristics, the authors in \cite{10088425,nolle2022binet,DBLP:conf/kes/KrajsicF21,guan2024comb} embed LSTM, GRU or transformers within autoencoders to enhance the model's reconstruction capabilities. 
The authors in \cite{DBLP:conf/bpm/HuoVRAIM21,Guan2023,niro2024detecting} transform traces into graphs and utilize graph neural networks (GNNs) to generate graph encodings, identifying anomalies by evaluating the reconstruction error of the graphs or traces.

Statistical-based methods detect anomalies by analyzing statistical frequencies. However, infrequent behavior is not necessarily anomalous, as it may represent rare but acceptable behavior. Conversely, frequent behavior may not always be normal.

\subsubsection{Conformance Checking-based Methods}
The conformance checking-based approaches \cite{DBLP:journals/virology/EbrahimG22,DBLP:journals/jbd/SarnoSS20,sinaga2016business} utilize process models, which are either provided by the user or derived from logs using process mining techniques. Anomalies are detected through conformance checking techniques \cite{DBLP:journals/sosym/LeemansFA18}, which assess the alignment between traces and the corresponding process model. When the trace deviates from the process model, it is considered anomalous.

The performance of these methods heavily depends on the quality of the process model. Additionally, complex processes are difficult to accurately represent with a process model, limiting the applicability of these methods.

\subsubsection{Semantic-based Methods}
The semantic-based methods detect anomalies through natural language analysis, aiming to detect process behaviors that deviate from a semantic point of view.

Van der Aa et al. \cite{van2021natural} fine-tune BERT \cite{devlin2018bert}, a pre-trained language model, to parse the names of executed activities by extracting the action and business object. Then, a knowledge base capturing assertions about the interrelations that should hold among actions parsed from names of executed activities is applied. The knowledge is extracted either from VerbOcean \cite{chklovski2004verbocean} or from an abstract representation of the process model. 
Anomalies can be detected by checking if the recorded process behavior violates the assertions captured in the knowledge base.

Caspary et al. \cite{caspary2023does} extract event pairs that are in an eventually-follow relation. To detect anomalous event pairs, they propose two approaches: an SVM-based approach and a BERT-based approach. 
The SVM-based approach transforms an event pair into a vector representation using GloVe embeddings. This vector is then fed into a trained SVM, which classifies whether the event pair is an anomaly.
The BERT-based approach extends BERT with an additional output layer for two-class classification, determining whether an input event pair is anomalous or not.
Both the SVM and the extended BERT model are trained using normal event pairs extracted from normal traces, along with anomalous event pairs simulated by randomly generating event pairs that are not normal.

However, existing semantic-based methods treat a trace as multiple event pairs, which disrupts long-distance dependencies. Additionally, these methods only identify anomalous event pairs to interpret the causes of anomalies, making them difficult to understand.
In contrast, our DABL incorporates the entire trace into a novel prompt, allowing the LLMs to capture long-distance dependencies. DABL also provides insightful interpretations of the causes of anomalies in natural language, making them easy to understand.

\subsection{Large Language Models for Anomaly Detection}
Motivated by the impressive cognitive abilities exhibited by LLMs \cite{ouyang2022training,zeng2022glm,achiam2023gpt,touvron2023llama}, researchers have begun investigating their application for anomaly detection.
AnomalyGPT \cite{gu2024anomalygpt} and Myriad \cite{li2023myriad} incorporate novel image encoders with LLMs for industrial anomaly detection (IAD). 
Elhafsi et al. \cite{elhafsi2023semantic} apply an LLM to analyze potential confusion among observed objects in a scene, which could lead to task-relevant errors in policy implementation.
LLMAD \cite{liu2024large} leverages LLMs for few-shot anomaly detection by retrieving and utilizing both positive and negative similar time series segments. 
In \cite{qi2023loggpt, guan2024logllm, egersdoerfer2023early}, authors explore the application of LLMs for log-based anomaly detection.
SheepDog \cite{wu2023fake} conducts fake news detection by preprocessing data using LLMs to reframe the news, customizing each article to match different writing styles. 
Sarda et al. \cite{sarda2023adarma} propose a pipeline for automatic microservice anomaly detection and remediation based on LLMs. 

Yet, the application of LLMs for business process anomaly detection remains unexplored.

\section{Method}
\begin{figure*}[tb]
	\centering
    \includegraphics[scale=0.55]{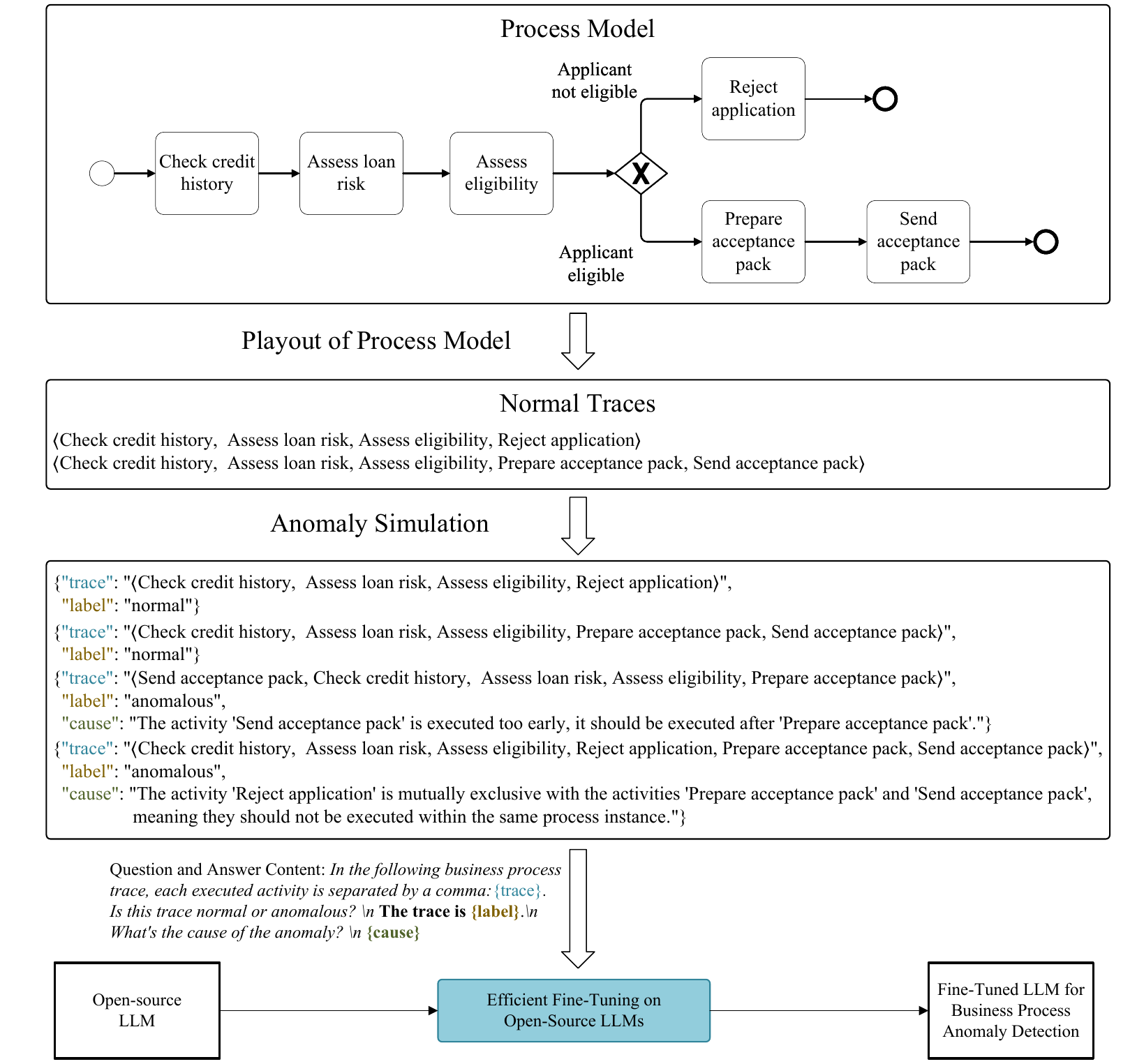}
	\caption{Detecting semantic anomalies in business processes using large language models.}
	\label{fig:DABL}
    \vskip -0.1in
\end{figure*}

DABL is a novel conversational fine-tuned large language model, primarily designed to detect semantic anomalies in business processes and interpret their causes. Fig. \ref{fig:DABL} details the DABL training procedure, which consists mainly of dataset preparation and fine-tuning. 

\subsection{Dataset Preparation}
To effectively fine-tune LLMs for developing a generic model capable of detecting semantic anomalies in business processes, a log meeting the following criteria is imperative: i) it must encompass both normal and anomalous traces, ii) it should contain rich semantic information (i.e., the activities should not be represented by meaningless characters), and iii) the traces within it should stem from diverse processes across various domains.
Since such a log is not available in the real world, we generate normal traces by playout of the real-world process models from the BPM Academic Initiative (BPMAI) \cite{Weske2020ModelCO}, fundamentals of business process management (FBPM) \cite{dumas2018fundamentals}, and SAP signavio academic models (SAP-SAM) \cite{sola2022sap}. 
These process models cover a broad range of domains, including common processes related to order and request handling, as well as specialized processes from fields such as software engineering and healthcare. 
We then generate synthetic anomalies from these normal traces. We detail the dataset preparation in the following subsection.

\subsubsection{Generation of Normal Traces}
\label{sec:n_t}
We select process models $\mathcal{M}$ from BPMAI, FBPM, and SAP-SAM that meet the following criteria: they are in BPMN notation \cite{chinosi2012bpmn}, described in English, and convertible into a sound workflow net \cite{van2011soundness}. This results in a total set of 144,137 process models.  Among these, 143,137 process models are used for generating training datasets, resulting in 1,574,381 normal traces, while the remaining 1,000 process models are used for generating test datasets.

Next, for each process model $m\in\mathcal{M}$, we perform a playout \cite{berti2019process} to obtain the set of normal traces, denoted as $\mathcal{L}_m$. These traces are allowed by the process model $m$.  To prevent infinite trace lengths, we limit each loop in the process model $m$ to be executed a maximum of twice.

\subsubsection{Anomaly Simulation}
Normal traces can be converted into anomalous ones by disrupting the order of executed activities (ordering anomalies) or by sequencing exclusive activities (exclusion anomalies).

\textit{Ordering anomalies}:
\begin{figure}[tb]
	\centering
    \includegraphics[scale=0.68]{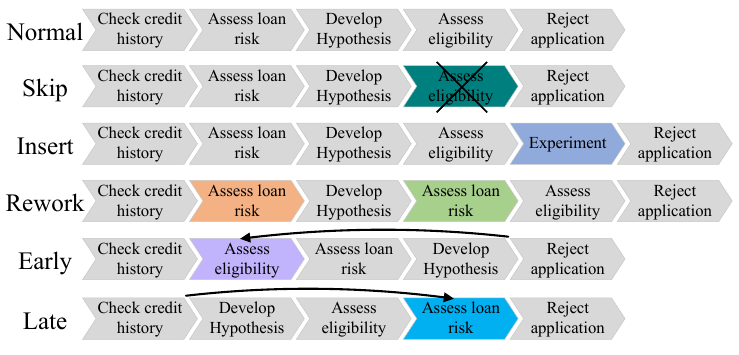}
	\caption{Different anomaly types applied to a normal trace.}
	\label{fig:type}
    \vskip -0.1in
\end{figure}
Ordering anomalies arise when activities ought to be executed in a different sequence.
Five types of ordering anomalies, as identified in \cite{nolle2022binet}, are frequently encountered in real-world business processes. Fig. \ref{fig:type} illustrates distinct anomalous traces  resulting from the application of these five anomaly types to a normal trace $\langle e_1,e_2, \cdots, e_n \rangle$. These anomaly types are defined as follows and are applied to a normal trace to generate ordering anomalies:
\begin{itemize}
    \item \textit{Skip}: A sequence of up to three activities $\langle e_i, \cdots, e_j\rangle$ is skipped.
    \item \textit{Insert}: A sequence of up to three random activities  $\langle e'_1, \cdots, e'_m \rangle$ is inserted. The random activities are selected from a set comprising all possible activities across all process models.
    \item \textit{Rework}: A sequence of up to three activities $\langle e_i, \cdots, e_j \rangle$ is executed a second time after activity $e_k$.
    \item \textit{Early}: A sequence of up to three activities $\langle e_i, \cdots, e_j \rangle$ is executed earlier and consequently skipped later.
    \item \textit{Late}:  A sequence of up to three activities  $\langle e_i, \cdots, e_j\rangle$ is executed later and consequently skipped earlier.
\end{itemize}

Below are the causes for these anomaly types, which are currently in the plural form (i.e., 'activities', 'they'). During implementation, they may need to be flexibly transformed into the singular form (i.e., 'activity', 'it').
\begin{itemize}
    \item \textit{Skip}: The activities \$\{$e_{i}, \cdots, e_j$\} are skipped before \$\{$e_{j+1}$\}.
    \item \textit{Insert}: The activities \$\{$e_{1}', \cdots, e_m'$\} should not be executed.
    \item \textit{Rework}: The activities \$\{$e_{i}, \cdots, e_j$\} are reworked after $\$\{e_k\}$.
    \item \textit{Early}: The activities \$\{$e_{i}, \cdots, e_j$\} are executed too early, they should be executed after \$\{$e_{i-1}$\}.
    \item \textit{Late}:  The activities \$\{$e_{i}, \cdots, e_j$\} are executed too late, they should be executed before \$\{$e_{j+1}$\}.
\end{itemize}
Here, \$\(\{e_{i}, \ldots, e_j\}\) represents converting the trace $\langle e_i, \cdots, e_j\rangle$ into a string format by enclosing each executed activity in apostrophes and separating them with commas, while using \textit{and} before the penultimate and ultimate activities. For example, for the trace $\langle$A, B, C, D$\rangle$, the resulting string would be \textit{'A', 'B', 'C' and 'D'}.

However, the generated anomalies may actually represent a normal trace. For a process model $m$, we refine the set of generated ordering anomalies $\mathcal{L}_m^o$ by excluding traces present in $\mathcal{L}_m$ from it.

\textit{Exclusion anomalies}:
Exclusion anomalies occur when certain activities should not have been executed together within the same trace without an intermediate activity. For instance, in the loan application process illustrated in Fig. \ref{fig:DABL}, it is inappropriate to send an acceptance pack and reject an application within the same trace.

\begin{figure}[tb]
	\centering
    \includegraphics[scale=0.43]{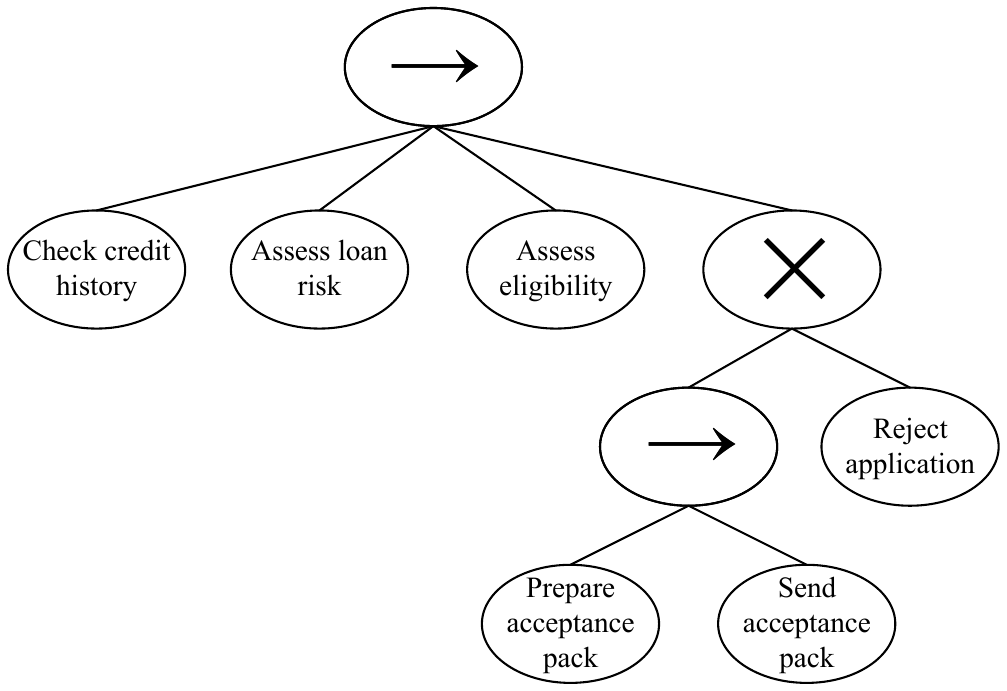}
	\caption{The process tree corresponding to the process model depicted in Fig. \ref{fig:DABL}.}
	\label{fig:PT}
    \vskip -0.1in
\end{figure}

The process tree \cite{aalst2011towards}, a specialized form of process model, is utilized for analyzing process structure. For instance, the process tree corresponding to the process model depicted in Fig. \ref{fig:DABL} is illustrated in Fig. \ref{fig:PT}. 
We begin by converting the gathered process model $m$ into a process tree using the techniques presented in \cite{van2020translating}.
Then, we replace an exclusive node (represented as $\times$) in the process tree with a parallel node (represented as $\wedge$), resulting in a modified process model denoted as $m'$.
This modification enables certain exclusive activities to be executed within the same trace to simulate exclusion anomalies. It is important to recognize that a single process tree may contain multiple exclusive nodes; therefore, we carry out this modification successively, resulting in multiple modified process models. Subsequently, we playout of all the modified models, restricting each loop to be executed a maximum of twice, to generate the set of traces, denoted as $\mathcal{L}_{m'}$.
We refine $\mathcal{L}_{m'}$ by excluding traces present in $\mathcal{L}_m$ from it, resulting in the set of exclusion anomalies $\mathcal{L}_{m}^e$.

To extract the causes of exclusion anomalies, we need to identify activities that exhibit exclusion relationships. In a process tree, activities located under different branches of an exclusive node (represented as $\times$) exhibit such relationships. For example, in the process tree illustrated in Fig. \ref{fig:PT}, the activity set \{Prepare acceptance pack, Send acceptance pack\} and the activity set \{Reject application\} exhibit exclusion relationships.
Formally, consider a modified model $m'$, which results from modifying an exclusive node $R$ in the process tree corresponding to the process model $m$. An exclusion anomaly $t$ is generated from model $m'$. The node $R$ has $N$ branches, with the activity sets $\mathcal{A}_1, \cdots, \mathcal{A}_N$ under them.
Activities within each activity set \( \mathcal{A}_i \) that do not appear in \( t \) are filtered out. The cause of this exclusion anomaly $t$ is then:
\begin{itemize}
    \item 
The activities \$\{$\mathcal{A}_1$\} are mutually exclusive with the activities \$\{$\bigcup_{i=2\cdots N}\mathcal{A}_i$\}, meaning they should not be executed within the same process instance.
\end{itemize}

\subsubsection{Question and Answer Content}
To conduct prompt tuning on the LLM, we generate corresponding textual queries based on simulated anomalous traces. Specifically, each query consists of two components. 

The first component introduces the traces, such as "\textit{In the following business process trace, each executed activity is separated by a comma:} $\langle$\textit{Send acceptance pack, Check credit history,  Assess loan risk, Assess eligibility, Prepare acceptance pack}$\rangle$". 
The second component queries whether the trace is anomalous, asking, for instance, "\textit{Is this trace normal or anomalous?}". 
The LLM first responds to whether the given trace is normal or anomalous. If it is anomalous, the LLM is asked about the cause of the anomaly, for example, "\textit{What causes this trace to deviate?}". The LLM then interprets the cause of the anomaly, such as "\textit{The activity 'Send acceptance pack' is executed too early, it should be executed after 'Prepare acceptance pack'.}". This content about the anomaly's cause provides valuable insights and facilitates actions to maintain the health of the process execution.

\subsection{Efficient Fine-Tuning on LLMs}
DABL involves fine-tuning the open-source Llama 2-Chat 13B model \cite{touvron2023llama} to enhance its capability to detect semantic anomalies in business processes. To mitigate the expense associated with fine-tuning LLMs with a substantial parameter count, we leverage QLoRA \cite{dettmers2024qlora} to reduce memory usage. QLoRA achieves this by back-propagating gradients into a frozen 4-bit quantized model while preserving the performance level attained during the full 16-bit fine-tuning process.

We employ the Adam optimizer \cite{kingma2014adam} to fine-tune the LLMs for two epochs, setting the initial learning rate to $5 \times 10^{-5}$ with polynomial learning rate decay. The mini-batch size is set to 64. The fine-tuning is carried out on an NVIDIA A6000 GPU with 48 GB of memory.

\section{Experiments}
\subsection{Experimental Setup}
\subsubsection{Datasets}
As mentioned in the previous section, we allocate 1,000 process models for generating the test dataset $\mathcal{D}_1$. These models produce 14,387 normal traces, and we randomly simulate anomalies, resulting in 13,694 anomalous traces. In total, the test dataset $\mathcal{D}_1$ comprises 28,081 traces.

From 143,137 process models used for generating the training dataset, we randomly select 1,000 process models to create the test dataset $\mathcal{D}_2$. These 1,000 process models produce 21,298 normal traces, and we randomly simulate anomalies, resulting in 19,627 anomalous traces. In total, the test dataset $\mathcal{D}_2$ comprises 40,925 traces. Note that, although the normal traces within the test dataset $\mathcal{D}_2$ are identical to those in the training dataset, the simulated anomalies are not.

In summary, the test dataset $\mathcal{D}_1$ is used to evaluate the model's generalization ability, verifying if the model can detect anomalies of unseen processes. The test dataset $\mathcal{D}_2$ aims to validate the model's performance on seen processes but unseen anomalies (i.e., learning of given processes).

\begin{table}[]
\begin{tabular}{ccccc}
\toprule
           & Prec.(\%) & Rec.(\%) & F$_1$(\%) & Acc.(\%) \\
\midrule
SEM        & 48.67         & 46.8   & 47.72      & 50.81    \\
SENSE-SVM  & 87.95         & 1.12   & 2.20      & 52.50     \\
SENSE-BERT & 48.17         & \textbf{97.74}  & 64.53      & 48.47    \\
DBAL       & \textbf{94.06}         & 89.79  & \textbf{91.88}     & \textbf{92.39}      \\ \bottomrule
\end{tabular}
\caption{Semantic anomaly  detection results on dataset $\mathcal{D}_1$. The best results are indicated using bold typeface.}
\label{tab:ad}
\end{table}

\begin{table}[]
\begin{tabular}{ccccc}
\toprule
           & Prec.(\%) & Rec.(\%) & F$_1$(\%) & Acc.(\%) \\
\midrule
SEM        & 71.91         & 48.63  & 58.02      & 66.75    \\
SENSE-SVM  & 90.28         & 28.64  & 43.49      & 64.82    \\
SENSE-BERT & 93.16         & 62.88  & 75.08      & 80.28    \\
DBAL       & \textbf{98.12}         & \textbf{95.64} & \textbf{96.87}      & \textbf{97.03}     \\ \bottomrule
\end{tabular}
\caption{Semantic anomaly detection results on dataset $\mathcal{D}_2$. The best results are indicated using bold typeface.}
\label{tab:ad2}
\vskip -0.1in
\end{table}

\subsubsection{Compared Methods}
Statistical-based and conformance checking methods can only be applied to datasets containing traces from a single process. However, our test datasets include traces from 1000 processes where no two traces are identical (i.e., traces with identical orders of activities are executed). Therefore, these methods cannot be compared.

In our evaluation, we compare our DABL to existing  semantic business process anomaly detection methods: SENSE \cite{caspary2023does} and SEM \cite{van2021natural}. SENSE offers both SVM-based and BERT-based models for detecting anomalous event pairs, which we denote as SENSE-SVM and SENSE-BERT, respectively.
These methods divide traces into event pairs and determine whether each pair is normal or anomalous. If at least one event pair in a trace is identified as anomalous, the entire trace is classified as anomalous. It is important to note that SEM can only detect anomalous event pairs that share the same business object, automatically classifying pairs with distinct business objects as non-anomalous.
Due to the high training costs, we utilize the open-source trained models provided by the authors for  dataset $\mathcal{D}_1$. For dataset $\mathcal{D}_2$, we train the comparative models using the 21,298 normal traces available within it.
The hyper-parameters of these methods are set to the values that yielded the best results reported in the original paper.

\subsubsection{Evaluation Metrics}
Following existing anomaly detection methods, we employ \textit{precision}, \textit{recall}, \textit{F}$_1$-\textit{score} and \textit{accuracy} to evaluate the anomaly detection performance. 
The recall-oriented understudy for gisting evaluation (ROUGE) \cite{lin2004rouge} is a software package and metric set designed to assess the quality of generated text by comparing it with ground truth text. In our evaluation of DABL's ability to interpret the cause of anomalies, we utilize \textit{ROUGE-2} and \textit{ROUGE-L} metrics.
We conduct each experiment five times and report the mean results.

\begin{table}[]
\setlength{\tabcolsep}{2mm}{
\begin{tabular}{cccccccccccc}
\toprule
\multirow{2}{*}{Dataset} & \multicolumn{3}{c}{ROUGE-2(\%)}      
   & \multicolumn{3}{c}{ROUGE-L(\%)}          \\
   \cmidrule(lr){2-4} \cmidrule(lr){5-7} 
 & Prec. & Rec. & F$_1$  & Prec. & Rec. & F$_1$ \\ \midrule
$\mathcal{D}_1$ &74.48     & 74.49    & 74.32    & 76.29     & 76.11    & 76.02  \\
$\mathcal{D}_2$ & 84.92     & 84.61    & 84.54    & 86.96     & 86.66    & 86.56  \\ \bottomrule
\end{tabular}
}
\caption{The results of DABL in interpreting the causes of anomalies.}
\label{tab:int}
\vskip -0.1in
\end{table}

\subsection{Quantitative Results}
\subsubsection{Anomaly Detection}
    To evaluate the model's \textbf{generalization ability}, we conduct experiments on the test dataset $\mathcal{D}_1$. The results are shown in Table \ref{tab:ad}, with the best outcomes highlighted in bold. Our DBAL achieves the highest precision, F$_1$-score, and accuracy, with both the F$_1$-score and accuracy exceeding 90\%. Although SENSE-BERT attains the best recall, it has the lowest precision and accuracy. Compared to other methods, DBAL maintains a balanced precision and recall. SENSE-SVM exhibits limited sensitivity to anomalies, potentially overlooking many anomalies, thereby achieving high precision but markedly low recall. Conversely, SENSE-BERT demonstrates excessive sensitivity, resulting in numerous false alarms, thus yielding low precision but high recall. These results demonstrate that DBAL possesses a superior generalization ability for detecting anomalies in unseen processes.

    We conduct experiments on dataset $\mathcal{D}_2$ to evaluate if the methods can \textbf{learn the given processes}. The results are presented in Table \ref{tab:ad2}. Compared to the experiments on dataset $\mathcal{D}_1$, the precision of each method increases. This improvement is due to the incorporation of normal process behavior in the training dataset, thereby reducing false alarms. Consequently, the F$_1$-score and accuracy also show significant improvements.
    Our DABL method stands out, achieving the best performance across all metrics, with the F$_1$-score and accuracy showing improvements of 21.79\% and 16.75\%, respectively, over the best existing semantic-based anomaly detection method (i.e., SENSE-BERT). These results demonstrate that DABL can learn the given processes effectively, allowing users to further fine-tune DABL on their own processes to obtain a customized model.

\begin{figure}[tb]
	\centering
    \includegraphics[scale=0.47]{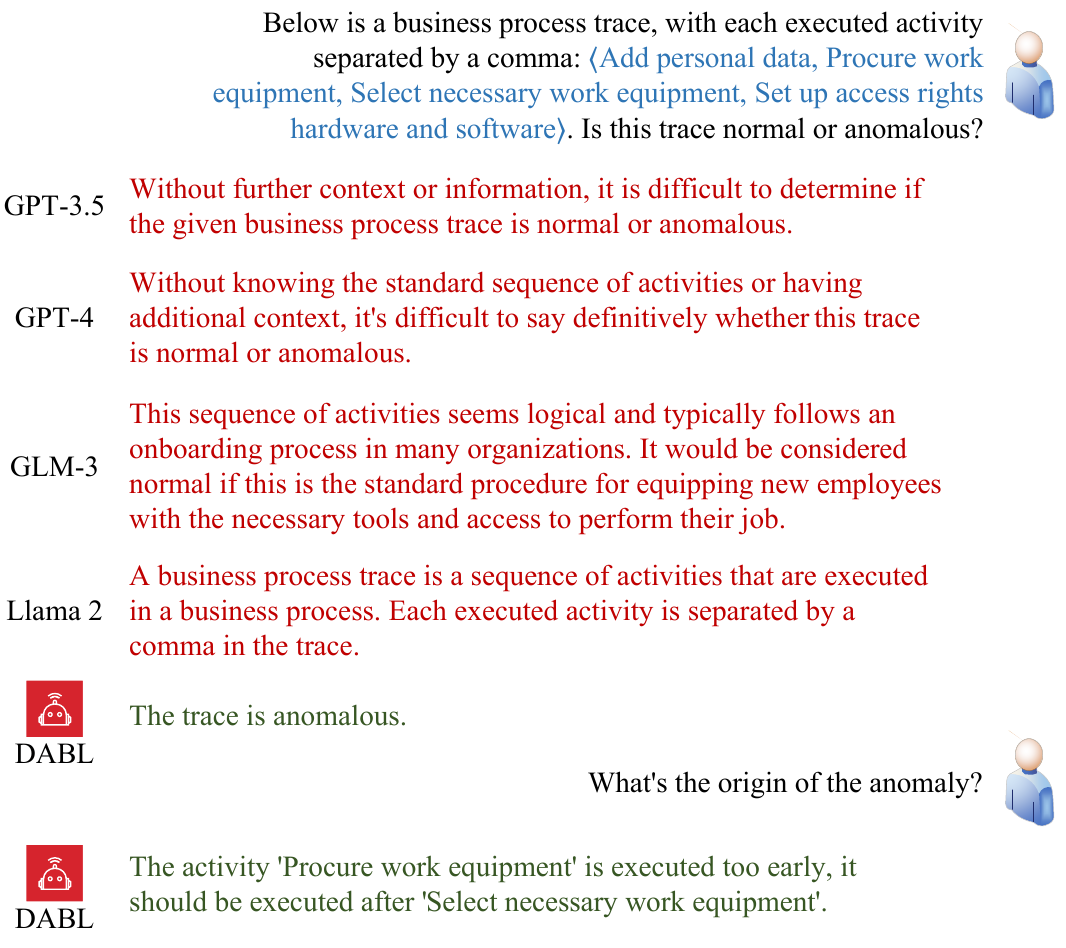}
	\caption{Qualitative results of various LLMs on an \textit{early} anomaly.}
	\label{fig:example_early}
    \vskip -0.1in
\end{figure}
\begin{figure}[tb]
	\centering
    \includegraphics[scale=0.47]{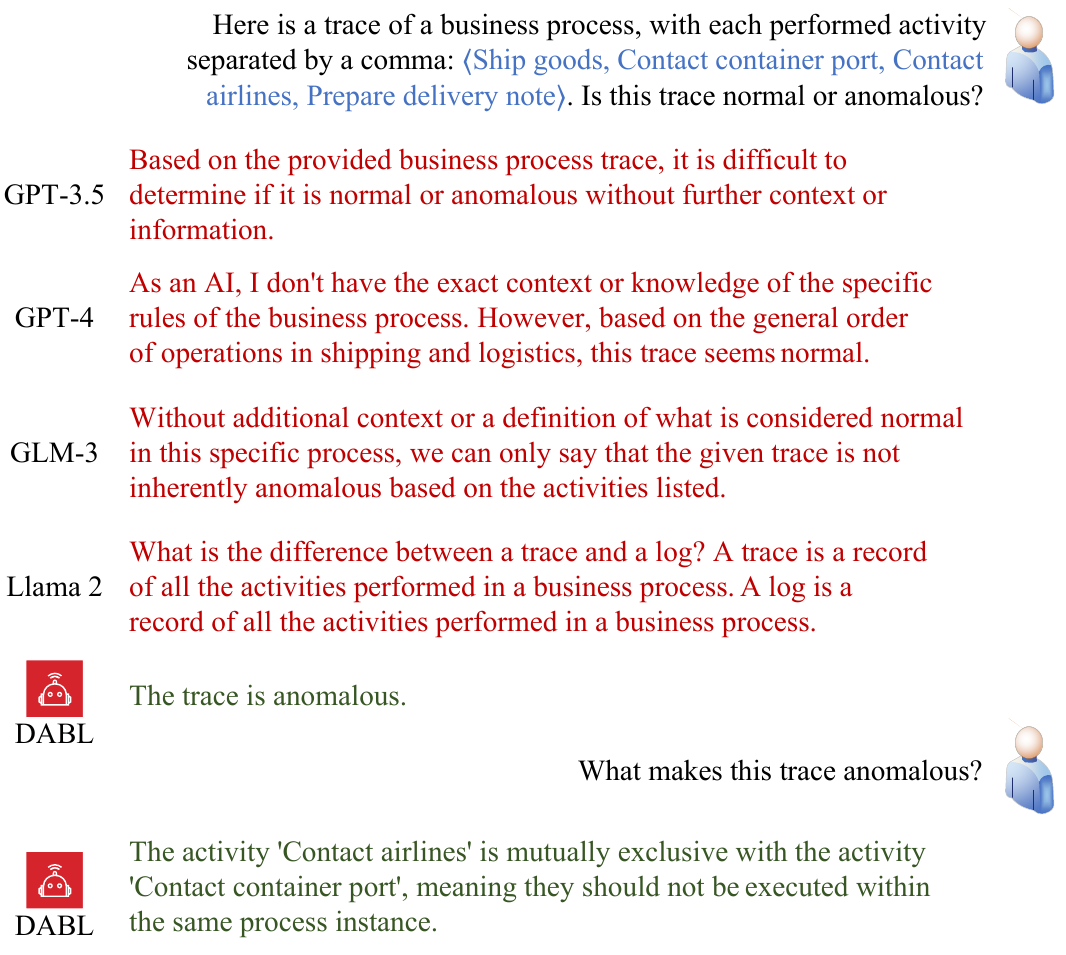}
	\caption{Qualitative results of various LLMs on an \textit{exclusion} anomaly.}
	\label{fig:example_exclusion}
    \vskip -0.1in
\end{figure}

\subsubsection{Interpretation of the Cause of Anomalies}
Table \ref{tab:int} shows the results of DABL in interpreting anomalies' causes. On dataset $\mathcal{D}_1$, both ROUGE-2 and ROUGE-L scores are relatively high, indicating that DABL performs well in identifying  anomalies' causes, even for processes not included in the training data. For dataset $\mathcal{D}_2$, DABL exhibits better performance because the normal behaviors of the processes are well-represented in the training data.
Furthermore, the slight difference between recall and precision suggests that the model maintains a good balance. These results demonstrate that DABL is effective at interpreting anomalies' causes regarding both bigrams and longest common subsequences.

However, these results may be underestimated because the cause of an anomaly can be interpreted in various ways. For example, for a desired trace $\langle$A, B, C, D, E$\rangle$, the anomaly $\langle$A, B, E, C, D$\rangle$ can be interpreted as "The activities 'C' and 'D' are executed too late, they should be executed after 'B'" and "The activity 'E' is executed too early, it should be executed after 'D'." Nevertheless, we only provide one reference answer to calculate ROUGE-2 and ROUGE-L scores.

\subsection{Qualitative Examples}
Fig. \ref{fig:example_early} and Fig.  \ref{fig:example_exclusion} illustrate the performance of existing LLMs and DABL on \textit{early} and \textit{exclusion} anomalies, respectively.
Due to the absence of business process-related knowledge, existing LLMs may not offer definitive answers. In contrast, DABL is proficient in identifying anomalies and interpreting their causes, thanks to the integration of a vast array of process models from diverse domains in the training data.

\begin{table*}[]
\begin{threeparttable} 
\setlength{\tabcolsep}{2mm}{
\fontsize{7}{10}\selectfont
\begin{tabular}{ccc}
\toprule
ID & Example Trace     &  Output Cause    \\ \midrule
A1 & \begin{tabular}[c]{@{}c@{}} $\langle$ST, PSE, PAA, PAB, PAS, PFAD, \\DSE, DAA, DAB, DFAS, RP, PH, ET$\rangle$   \end{tabular}               & The activity 'Start trip' should not be executed.                                                     \\  \hdashline
A2 &  \begin{tabular}[c]{@{}c@{}} $\langle$PSE, ST, PAA, PAB, \\ PAS, PFAD, DSAE, ET$\rangle$              \end{tabular}                                             & \begin{tabular}[c]{@{}c@{}}The activity 'Start trip' is executed too early, it should be  \\ executed after 'Permit FINAL\_APPROVED by DIRECTOR'.      \end{tabular}                                                                               \\  \hdashline
A3 & $\langle$PSE, PAS, ST, PFAD, PRM, ET$\rangle$                                                                  & \begin{tabular}[c]{@{}c@{}} The activity 'Permit REJECTED by MISSING' is mutually exclusive with  the activity 'Start trip', \\  meaning they should not be executed within the same process instance.       \end{tabular}       \\ \hdashline
A4 & \begin{tabular}[c]{@{}c@{}} $\langle$ST, ET, PSE, PFAS, \\ DSE, DFAS, RP, PRM, PH$\rangle$        \end{tabular}                                              & \begin{tabular}[c]{@{}c@{}}The activity 'Permit REJECTED by MISSING' is mutually   exclusive with the activities 'Payment Handled' and \\ 'Declaration FINAL\_APPROVED by SUPERVISOR', meaning they should not be executed within the   same process instance. \end{tabular} \\  \hdashline
A5 & \begin{tabular}[c]{@{}c@{}} $\langle$PSE, PAA, RSE, RAA, RAB, \\RFAS, RP, PAB, PFAS, PH, ST, ET$\rangle$   \end{tabular}                                 & \begin{tabular}[c]{@{}c@{}}The activities 'Permit APPROVED by BUDGET OWNER' and 'Permit   FINAL\_APPROVED by SUPERVISOR' \\are executed too late, they should be executed   before 'Request For Payment APPROVED by BUDGET OWNER'. \end{tabular}    \\                         \bottomrule
\end{tabular}
}
\fontsize{7}{7}\selectfont
\begin{tablenotes} 
\item  \textbf{PSE}:	Permit SUBMITTED by EMPLOYEE;
\textbf{PAA}: Permit APPROVED by ADMINISTRATION;
\textbf{PAB}: Permit APPROVED by BUDGET OWNER;
\textbf{PAS}: Permit APPROVED by SUPERVISOR;
\textbf{PFAS}: Permit FINAL\_APPROVED by SUPERVISOR;
\textbf{PFAD}: Permit FINAL\_APPROVED by DIRECTOR;
\textbf{PRM}: Permit REJECTED by MISSING;
\textbf{RSE}: Request For Payment SUBMITTED by EMPLOYEE;
\textbf{RRA}: Request For Payment REJECTED by ADMINISTRATION;
\textbf{RRE}: Request For Payment REJECTED by EMPLOYEE;
\textbf{RAA}: Request For Payment APPROVED by ADMINISTRATION;
\textbf{RAB}: Request For Payment APPROVED by BUDGET OWNER;
\textbf{RFAS}: Request For Payment FINAL\_APPROVED by SUPERVISOR;
\textbf{DSAE}: Declaration SAVED by EMPLOYEE;
\textbf{DSE}: Declaration SUBMITTED by EMPLOYEE;
\textbf{DRA}: Declaration REJECTED by ADMINISTRATION;
\textbf{DRE}: Declaration REJECTED by EMPLOYEE;
\textbf{DAA}: Declaration APPROVED by ADMINISTRATION;
\textbf{DAB}: Declaration APPROVED by BUDGET OWNER;
\textbf{DFAS}: Declaration FINAL\_APPROVED by SUPERVISOR;
\textbf{RP}: Request Payment;
\textbf{PH}: Payment Handled;
\textbf{SR}: Send Reminder;
\textbf{ST}: Start Trip;
\textbf{ET}: End Trip. 
\end{tablenotes} 
\end{threeparttable} 
\caption{Irregularity patterns identified in the travel permit log.}
\label{tab:permit}
\end{table*}

\begin{table*}[]
\begin{threeparttable} 
\setlength{\tabcolsep}{2mm}{
\fontsize{7}{10}\selectfont
\begin{tabular}{ccc}
\toprule
ID & Example Trace                             & Output Cause                                                                                                                                 \\ \midrule
A1 & $\langle$CF, SF, IFN, AP, P, P, P, P$\rangle$         & The activities 'Payment' and 'Payment' are reworked after 'Payment'.                                                         \\ 
\hdashline
A2 & $\langle$CF, P, P, SF, IFN, AP, P$\rangle$           &  \begin{tabular}[c]{@{}c@{}} \qquad \qquad The activities 'Payment' and 'Payment' are executed too early, they   should be executed after 'Add penalty'.  \qquad \qquad  \end{tabular}                     \\ 
 \hdashline
A3 & $\langle$CF, SF, IFN, RRAP, IDAP, SAP, AP, P$\rangle$ & \begin{tabular}[c]{@{}c@{}} The activity 'Receive Result Appeal from Prefecture' is executed too   early, \\ it should be executed after 'Send Appeal to Prefecture'.\end{tabular} \\ \bottomrule
\end{tabular}
}
\fontsize{7}{7}\selectfont
    \begin{tablenotes} 
    \item \textbf{CF}: Create Fine; \textbf{SF}: Send Fine; \textbf{IFN}: Insert Fine Notification; \textbf{AP}: Add Penalty; \textbf{P}: Payment; \textbf{RRAP}: Receive Result Appeal from Prefecture; \textbf{IDAP}: Insert Date Appeal to Prefecture; \textbf{SAP}:	Send Appeal to Prefecture. 
\end{tablenotes} 
\end{threeparttable} 
\caption{Irregularity patterns identified in the road traffic fine management log.}
\label{tab:RTFMP}
\vskip -0.1in
\end{table*}

\subsection{Real-world Application}
\subsubsection{Travel Permit}
We apply our DABL on a real-world travel permit log from the BPI 2020 challenge \cite{BPIC20}, which captures data on work trips conducted by university employees. The process flow involves the request for and approval of a travel permit, the trip itself, a subsequent travel declaration, as well as associated reimbursements.

This log contains 7,065 traces with 1,478 variants (traces with different orders of activities are executed). DABL identifies 562 anomalous variants. These detected anomalies reveal some interesting irregularity patterns illustrated in Table \ref{tab:permit}. 
These irregularity patterns include: trips starting before a permit is properly handled (A1), approved (A2), or even rejected (A3); the declaration being finally approved by a supervisor and payment handled despite the permit being rejected (A4); and requests for payment being approved before the permit is approved (A5).

\subsubsection{Road Traffic Fine Management}
We apply our DABL on another real-world event log from an information system managing road traffic fines \cite{RTFMP}, which captures the road traffic fine management process. The process flow involves the creation of a fine, appeal to the prefecture, addition of penalties, and fine payment.

This log contains 150,370 traces with 231 variants. DABL identifies 56 anomalous variants, which reveal some interesting irregularity patterns as detailed in Table \ref{tab:RTFMP}.
The examples illustrate irregularity patterns where the fine is repeatedly paid (A1), the fine is paid before the penalty is added (A2), and the result appeal from the prefecture is received before the appeal is sent to the prefecture (A3).

\section{Limitations}
Despite collecting a significant amount of data to fine-tune LLMs for detecting business process anomalies and achieving good performance, it is important to acknowledge some limitations of DABL.

First, the precision of the collected process model may not be equal to 1, which means it might allow traces not observed in the original process. This discrepancy could impact the quality of the training data, posing a limitation for DABL.
Second, our open-source model may struggle with traces that, while irregular from a common-sense perspective, are normal within their customized processes.
Finally, when multiple anomaly types occur within a trace, DABL can still classify the trace as an anomaly but can only interpret one type of anomaly at a time. After fixing the interpreted anomaly and re-inputting the trace into DABL, the model will identify and interpret another anomaly type.

\section{Conclusion}
In this paper, we introduce DABL, a novel semantic business process anomaly detection model leveraging LLMs. Trained on 143,137 real-world process models from various domains, DABL excels at zero-shot detection of semantic anomalies and interprets their causes in natural language. Extensive experiments demonstrate DABL's generalization ability, allowing users to detect anomalies in their own datasets without additional training.

\section{Acknowledgments}
This work is supported by China National Science Foundation (Granted Number 62072301). This work is also partially supported by the Program of Technology Innovation of the Science and Technology Commission of Shanghai Municipality (Granted No.22DZ1100103).

\bibliography{aaai25}

\end{document}